\documentclass{article}

\usepackage[square,numbers]{natbib}

     \usepackage[final]{neurips_data_2021}

\usepackage[utf8]{inputenc} 
\usepackage[T1]{fontenc}    
\usepackage{hyperref}       
\usepackage{url}            
\usepackage{booktabs}       
\usepackage{amsfonts}       
\usepackage{nicefrac}       
\usepackage{microtype}      
\usepackage{xcolor}         

\usepackage{xcolor}
\usepackage{multicol}
\usepackage{multirow}
\usepackage{makecell}
\usepackage{graphicx}
\usepackage{booktabs}
\usepackage{colortbl}
\usepackage{xspace}
\usepackage{amssymb}
\usepackage{pifont}
\usepackage{paralist}
\usepackage{subfig}

\newcommand\sect[1]{\S\ref{#1}}
\newcommand\cose{\textsc{\small COS-E}\xspace}
\newcommand\exnlp{\textsc{\small ExNLP}\xspace}
\newcommand\esnli{\textsc{\small E-SNLI}\xspace}
\newcommand\esnlivte{\textsc{\small E-SNLI-VE}\xspace}
\newcommand\snlive{\textsc{\small SNLI-VE}\xspace}
\newcommand\snli{\textsc{\small SNLI}\xspace}
\newcommand\mnli{\textsc{\small MNLI}\xspace}

\newcommand\flickr{\textsc{\small Flickr30k}\xspace}
\newcommand\vqae{\textsc{\small VQA-E}\xspace}
\newcommand\edit{\textsc{\small EDIT}\xspace}
\newcommand\judge{\textsc{\small JUDGE}\xspace}
\newcommand\glucose{\textsc{\small GLUCOSE}\xspace}
\newcommand\CollectAndEdit{\textsc{\small Collect-And-Edit}\xspace}
\newcommand\CollectAndJudge{\textsc{\small Collect-And-Judge}\xspace}

\definecolor{coolpink}{RGB}{252, 8, 142}

\title{Teach Me to Explain: A Review of Datasets for Explainable Natural Language Processing}

\author{Sarah Wiegreffe\thanks{\enspace Equal contributions.} \\
  School of Interactive Computing \\
  Georgia Institute of Technology \\
  {\tt saw@gatech.edu} \\\And
  Ana Marasovi\'{c}\footnotemark[1] \\
  Allen Institute for AI \\
  University of Washington\\
  {\tt anam@allenai.org}\\}

\begin{document}

\maketitle

\begin{abstract}
Explainable Natural Language Processing (\exnlp) has increasingly focused on collecting human-annotated textual explanations. These explanations are used downstream in three ways: as data augmentation to improve performance on a predictive task, as supervision to train models to produce explanations for their predictions, and as a ground-truth to evaluate model-generated explanations. In this review, we identify 65 datasets with three predominant classes of textual explanations (highlights, free-text, and structured), organize the literature on annotating each type, identify strengths and shortcomings of existing collection methodologies, and give recommendations for collecting \exnlp datasets in the future.
\end{abstract}

\section{Introduction}
\label{sec:introduction}

Interpreting supervised machine learning (ML) models is crucial for ensuring their reliability and trustworthiness in high-stakes scenarios. 
Models that produce justifications for their individual predictions (sometimes referred to as \textit{local explanations}) can be inspected for the purposes of debugging, quantifying bias and fairness, understanding model behavior, and ascertaining robustness and privacy \cite{molnar2019}. 
These benefits have led to the development of 
datasets that contain human justifications for the true label
(overviewed in Tables \ref{table:highlights_overview}--\ref{table:structured_overview}). In particular, human justifications are used for three goals: (i) to aid models with additional training supervision 
\cite{zaidan-etal-2007-using}, 
(ii) to train interpretable models that explain their own predictions \cite{camburu2018snli},  
and (iii) to evaluate plausibility of 
model-generated explanations by measuring their agreement with human 
explanations \cite{deyoung-etal-2020-eraser}. 


Dataset collection is the most under-scrutinized component of the ML pipeline \cite{paritosh2020}---it is estimated that 92$\%$ of ML practitioners encounter data cascades, or downstream problems resulting from poor data quality \cite{sambasivan2020}. It is important to constantly evaluate data collection practices critically and 
standardize them \cite{bender-friedman-2018-data, Gebru2018DatasheetsFD, Paullada2020DataAI}. 
We expect that such examinations are  particularly valuable when many related datasets are released contemporaneously and independently in a short period of time, as is the case with \exnlp datasets.  

This survey aims to review and summarize the literature on collecting textual explanations, 
highlight what has been learned to date, and give recommendations for future 
dataset construction. %
It complements other explainable AI (\textsc{\small XAI}) surveys and critical retrospectives that focus on definitions, methods, and/or evaluation 
\cite{doshi2017towards, biran2017explanation, lipton2018mythos, adadi-berrada-peeking-2018, Ras2018, hoffman2018metrics, Gilpin2018ExplainingEA, Yang2019EvaluatingEW, clinciu-hastie-2019-survey, Guidotti2019ASO, miller2019explanation, Verma2020CounterfactualEF, BARREDOARRIETA202082, Murdoch22071, jacovi-goldberg-2020-towards, Burkart2021ASO}, but not on datasets. We call such datasets \exnlp datasets, because modeling them for the three goals mentioned above requires NLP techniques.
Datasets and methods for explaining 
fact checking \cite{kotonya-toni-2020-survey}  and 
reading comprehension \cite{Thayaparan2020ASO} have been reviewed; we are the first to review all datasets with textual explanations regardless of task, comprehensively categorize them into three distinct classes, and provide critical retrospectives and best-practice recommendations.

We first define relevant \exnlp terminology (\sect{sec:terminology}) and overview 65 existing datasets (\sect{sec:survey_datasets}), accompanied with a live version of the tables as a website accepting community contributions: \url{https://exnlpdatasets.github.io}. We next analyze what can be learned from existing data collection methodologies. In  \sect{sec:collecting_highlights} and \sect{sec:case_study_cose_esnli}, we highlight two points that we expect to be particularly important to the current ExNLP research. Specifically, \sect{sec:collecting_highlights} discusses the traditional process of collecting explanations by asking annotators 
to highlight parts of the input, and its discrepancies 
with evaluating model-generated highlight explanations. 
We also draw attention to how assumptions made for collecting free-text explanations (introduced in \sect{sec:terminology}) influence their modeling, and call for better documentation of explanation collection.  
In \sect{sec:case_study_cose_esnli}, we illustrate that not all template-like free-text explanations are incorrect, and call for embracing the structure of an explanation when appropriate. 
Unlike discussions in \sect{sec:collecting_highlights}--\ref{sec:case_study_cose_esnli} that are motivated by \exnlp modeling and evaluation choices, the rest of this paper reflects on relevant points from a broader NLP research. In \sect{sec:quality}, we present a proposal for controlling quality in explanation collection, and in  \sect{sec:related_quality}, gather recommendations from related subfields to further reduce data artifacts by increasing diversity of collected explanations. 

\begin{table*}[t]
    \centering
    \resizebox{\textwidth}{!}{
\begin{tabular}{p{9.46cm}p{9.46cm}}
\toprule
\textbf{Instance} & \textbf{Explanation} \\
\midrule 
\multirow{4}{9.46cm}{\emph{Premise:} A white race dog wearing the number eight runs on the track. \hspace{4in} \emph{Hypothesis:} A white race dog runs around his yard. \hspace{4in} \emph{Label:} contradiction} &
\textbf{\texttt{{\color{coolpink}(highlight)}}} \emph{Premise:} A white race dog wearing the number eight runs on the \colorbox{yellow!30}{track}. \emph{Hypothesis:} A white race dog runs around his \colorbox{yellow!30}{yard}.  \\
\arrayrulecolor{black!20}\cmidrule{2-2}
 & \textbf{\texttt{{\color{coolpink}(free-text)}}} A race track is not usually in someone's yard.  \\
\arrayrulecolor{black!20}\midrule
\emph{Question:}  Who sang the theme song from Russia With Love? \hspace{4in} \emph{Paragraph:} \dots The theme song was composed by Lionel Bart of Oliver! fame and sung by Matt Monro\dots \hspace{4in} \emph{Answer:} Matt Monro  &  \textbf{\texttt{{\color{coolpink}(structured)}}} \emph{Sentence selection:} (not shown) \hspace{1in} \emph{Referential equality:} ``the theme song from russia with love'' (from question) = ``The theme song'' (from paragraph) \hspace{2in} \emph{Entailment:} X was composed by Lionel Bart of Oliver! fame and sung by ANSWER. $\vdash$ ANSWER sung X \\

\arrayrulecolor{black}\bottomrule 
\end{tabular}}
\caption{Examples of explanation types discussed in \sect{sec:terminology}. The first two rows show a highlight and free-text explanation for an \esnli instance \cite{camburu2018snli}. The last row shows a (partial) structured explanation from \textsc{\small QED} for a \textsc{\small NaturalQuestions} instance \cite{lamm2020qed}.}
\label{tab:explanation_examples}
\end{table*}

\section{Explainability Lexicon} 
\label{sec:terminology}

An explanation can be described as a ``three-place predicate: \emph{someone} explains \emph{something} to \emph{someone}'' \cite{hilton1990conversational}. The \emph{something} being explained in machine learning systems are task labels: explanations are implicitly or explicitly designed to answer the question ``why is [input] assigned [label]?''. However, collected explanations can vary in format.
We identify three types in the \exnlp literature: \emph{highlights}, \emph{free-text}, and \emph{structured} explanations. An example of each type is given in \autoref{tab:explanation_examples}. Since a consensus on terminology has not yet been reached, we describe each type below. 

\begin{table*}[t]
    \centering
    \resizebox{\textwidth}{!}{
\begin{tabular}{p{9cm}p{9.12cm}}
\toprule
\textbf{Instance with Highlight} & \textbf{Highlight Type Clarification} \\
\midrule 
\emph{Review:} this film is \colorbox{yellow!30}{extraordinarily horrendous} and I'm not going to waste any more words on it. \emph{Label}: negative & \textbf{\texttt{{\color{coolpink}($\lnot$comprehensive)}}} \emph{Review:} this film is  \colorbox{black}{extraordinari horrend} and I'm not going to waste any more words on it. \\
\arrayrulecolor{black!20}\midrule
\emph{Review:} this film is \colorbox{yellow!30}{extraordinarily horrendous} and I'm not going to \colorbox{yellow!30}{waste any more words on it}. \emph{Label}: negative & \textbf{\texttt{{\color{coolpink}(comprehensive)}}} \emph{Review:} this film is  \colorbox{black}{extraordinari horrend} and I'm not going to \colorbox{black}{waste any more words on i}. \\
\arrayrulecolor{black!20}\midrule
\emph{Premise:} A shirtless man wearing white shorts. \emph{Hypothesis:} A \colorbox{yellow!30}{man} in white shorts is \colorbox{yellow!30}{running on the sidewalk.} \emph{Label}: neutral &  \textbf{\texttt{{\color{coolpink}($
\lnot$sufficient)}}} \emph{Premise:} \colorbox{black}{A shirtless man wearing xxxxx} \emph{Hypothesis:} \colorbox{black}{A} \colorbox{yellow!30}{man} \colorbox{black}{in white shorts} \colorbox{yellow!30}{running on the sidewalk.}  \\
\arrayrulecolor{black}\bottomrule 
\end{tabular}}
\caption{Examples of highlights differing in comprehensiveness and sufficiency (discussed in \sect{sec:terminology}, \sect{sec:collecting_highlights}).}
\label{tab:explanation_examples_two}
\end{table*}


\begin{table*}[t]
    \centering
    \resizebox{\textwidth}{!}{
    \begin{tabular}{llllrrr}
    \toprule
        \textbf{Dataset} & \textbf{Task} & \textbf{Granularity} & \textbf{Collection} & \textbf{\# Instances}\\
        \midrule
        \textsc{\small MovieReviews} \cite{zaidan-etal-2007-using} & sentiment classification & none & author & 1,800 \\
        \textsc{\small MovieReviews$_{c}$} \cite{deyoung-etal-2020-eraser} & sentiment classification & none & crowd  & 200$^{\ddagger\diamondsuit}$\\
        \textsc{\small SST} \cite{socher-etal-2013-recursive} & sentiment classification  & none & crowd &  11,855$^{\diamondsuit}$\\
        \textsc{\small WikiQA} \cite{yang-etal-2015-wikiqa} & open-domain QA & sentence & crowd + authors &  1,473\\ 
        \textsc{WikiAttack} \cite{carton-etal-2018-extractive} & detecting personal attacks &  none  & students &  1089$^{\diamondsuit}$\\
        \textsc{\small E-SNLI}$^{\dagger}$ \cite{camburu2018snli} & natural language inference & none & crowd &  $\sim$569K (1 or 3) \\ 
        \textsc{\small MultiRC} \cite{khashabi-etal-2018-looking} & reading comprehension QA & sentences & crowd &  5,825 \\
        \textsc{\small FEVER} \cite{thorne-etal-2018-fever} &  verifying claims from text & sentences & crowd 
        &  $\sim$136K$^{\ddagger}$ \\
        \textsc{\small HotpotQA} \cite{yang-etal-2018-hotpotqa} &  reading comprehension QA & sentences & crowd &  112,779 \\
        \citet{hanselowski-etal-2019-richly} &  verifying claims from text & sentences & crowd &  6,422 (varies) \\
        \textsc{\small NaturalQuestions} \cite{kwiatkowski-etal-2019-natural} &  reading comprehension QA & 1 paragraph & crowd & n/a$^{\ddagger}$ (1 or 5) \\
        \textsc{\small CoQA} \cite{reddy-etal-2019-coqa} & conversational QA & none & crowd & $\sim$127K (1 or 3)\\
         \cose \textsc{\small v1.0}$^{\dagger}$ \cite{rajani-etal-2019-explain} &  commonsense QA  & none & crowd &  8,560 \\
        \cose \textsc{\small v1.11}$^{\dagger}$ \cite{rajani-etal-2019-explain}  &  commonsense QA  & none & crowd &  10,962 \\
        \textsc{\small BoolQ$_{c}$}
        \cite{deyoung-etal-2020-eraser}  & reading comprehension QA  & none & crowd &  199$^{\ddagger\diamondsuit}$ \\
        \textsc{\small EvidenceInference v1.0} \cite{lehman-etal-2019-inferring} & evidence inference & none & experts  &  10,137 \\
        \textsc{\small EvidenceInference v1.0$_{c}$} \cite{deyoung-etal-2020-eraser} & evidence inference & none & experts &  125$^{\ddagger}$ \\
        \textsc{\small EvidenceInference v2.0} \cite{deyoung-etal-2020-evidence} &  evidence inference & none & experts  &  2,503 \\
        \textsc{\small SciFact} \cite{wadden-etal-2020-fact} & verifying claims from text  &  1-3 sentences & experts & 995$^{\ddagger}$ (1-3)\\
         \citet{kutlu2020annotator} & webpage relevance ranking & 2-3 sentences & crowd & 700 (15) \\
         \textsc{\small SCAT} \cite{Yin2021DoCT} & document-level machine translation & none & experts & $\sim$14K\\
         \textsc{\small ECtHR} \cite{chalkidis-etal-2021-paragraph} & alleged legal violation prediction & paragraphs & auto + expert & $\sim$11K\\
         \textsc{Hummingbird} \cite{Hayati2021DoesBL} & style classification & words & crowd & 500 \\
         \textsc{HateXplain} \cite{Mathew2021HateXplainAB} & hate-speech classification & phrases & crowd & 20,148 (3) \\
        \bottomrule
    \end{tabular}}
    \caption{Overview of datasets with textual \textbf{highlights}. Values in parentheses indicate number of explanations collected per instance (if $>$ 1). \citet{deyoung-etal-2020-eraser} collected or recollected annotations for prior datasets (marked with the subscript  ${c}$). $\diamondsuit$ Collected $>$ 1 explanation per instance but only release 1. $\dagger$ Also contains free-text explanations. $\ddagger$ A subset of the original dataset that is annotated. It is not reported what subset of \textsc{\small NaturalQuestions} has both a long and short answer.}
    \label{table:highlights_overview}
\end{table*}

\textbf{Highlights} are subsets of the input elements (words, phrases, or  sentences) that explain a prediction.
\citet{lei-etal-2016-rationalizing} 
coin them \emph{extractive rationales}, or subsets of the input tokens of a textual task that satisfy two properties: (i) \emph{compactness}, 
they are short and coherent, and (ii) \emph{sufficiency}, 
they suffice for prediction as a substitute of the original text. 
\citet{yu-etal-2019-rethinking} introduce a third criterion, (iii) \emph{comprehensiveness}, that all the evidence that supports the prediction is selected, not just a sufficient set.
Since the term ``rationale'' implies human-like intent, \citet{jacovi2020aligning} argue to call this type of explanation \emph{highlights} 
to avoid inaccurately attributing human-like social behavior to AI systems. They are also called \emph{evidence} in fact-checking and multi-document question answering (QA) \cite{kotonya-toni-2020-survey}---a part of the source that refutes/supports the claim.  
To reiterate, highlights should be sufficient to explain a prediction and compact; if they are also comprehensive, we call them \emph{comprehensive highlights}. Although the community has settled on criteria (i)--(iii) for highlights, the extent to which collected datasets (Table \ref{table:highlights_overview}) reflect them varies greatly, as we will discuss in \sect{sec:collecting_highlights}. Table \ref{tab:explanation_examples_two} gives examples of sufficient vs.\ non-sufficient and comprehensive vs.\ non-comprehensive highlights.

\textbf{Free-text explanations} are free-form textual justifications that are not constrained to the words or modality of the input instance.  
They are thus more expressive and generally more readable than highlights. This makes them useful for explaining reasoning tasks where explanations must contain information outside the given input sentence or document \cite{camburu2018snli, Wiegreffe2020MeasuringAB}. 
They are also called \emph{textual} \cite{Kim2018TextualEF} or \emph{natural language explanations} 
\cite{camburu2018snli}, terms that have been overloaded \cite{Prasad2020ToWE}.
Synonyms, \emph{free-form} \cite{camburu2018snli} or \emph{abstractive explanations} \cite{narang2020wt5} 
do not emphasize that the explanation is textual.

Finally, \textbf{structured explanations} are
explanations that are not entirely free-form although they are still written in natural language. For example, there may be constraints placed on the explanation-writing process, such as the required use of specific inference rules. We discuss the recent emergence of structured explanations in \sect{sec:case_study_cose_esnli}. Structured explanations do not have one common definition; we elaborate on dataset-specific designs in \sect{sec:survey_datasets}. An example is given in the bottom row of \autoref{tab:explanation_examples}.

\begin{table*}[t]
    \centering
    \resizebox{0.89\textwidth}{!}{
      \begin{tabular}{lllrrr}
    \toprule
        \textbf{Dataset} & \textbf{Task} & \textbf{Collection} & \textbf{\# Instances} \\        \midrule
        \citet{jansen-etal-2016-whats} & science exam QA & authors & 363\\
        \citet{ling-etal-2017-program} & solving algebraic word problems  & auto + crowd &  $\sim$101K\\
        \citet{srivastava-etal-2017-joint}$^\ast$ & detecting phishing emails  & crowd + authors &  7 (30-35)\\
        \textsc{\small BabbleLabble} \cite{hancock-etal-2018-training}$^\ast$ & relation extraction  & students + authors &  200$^{\ddagger\ddagger}$
        \\
        \esnli \cite{camburu2018snli} & natural language inference & crowd &  $\sim$569K (1 or 3) \\
        \textsc{\small LIAR-PLUS} \cite{alhindi-etal-2018-evidence} &  verifying claims from text  & auto & 12,836 \\
        \cose \textsc{\small v1.0} \cite{rajani-etal-2019-explain} & commonsense QA &  crowd & 8,560 \\
        \cose \textsc{\small v1.11} \cite{rajani-etal-2019-explain}  & commonsense QA   & crowd & 10,962 \\
        \textsc{\small ECQA} \cite{aggarwal-etal-2021-explanations} & commonsense QA & crowd & 10,962 \\
        \textsc{\small Sen-Making} \cite{wang-etal-2019-make} & commonsense validation & students + authors & 2,021 \\
        \textsc{\small ChangeMyView} \cite{atkinson-etal-2019-gets} & argument persuasiveness & crowd & 37,718 \\
        \textsc{\small WinoWhy} \cite{zhang-etal-2020-winowhy} & pronoun coreference resolution  & crowd  & 273 (5)\\
        \textsc{\small SBIC} \cite{sap-etal-2020-social} & social bias inference & crowd & 48,923 (1-3) \\ 
        \textsc{\small PubHealth} \cite{kotonya-toni-2020-explainable-automated} & verifying claims from text   & auto  &  11,832 \\
        \citet{wang2019learning}$^\ast$ & relation extraction  & crowd + authors &  373  \\
         \citet{wang2019learning}$^\ast$ & sentiment classification  & crowd + authors &  85  \\
        \textsc{\small E-$\delta$-NLI} \cite{Brahman2020LearningTR} & defeasible natural language inference & auto & 92,298  ($\sim$8) \\
        \arrayrulecolor{black!20}\midrule
        \textsc{\small BDD-X}$^{\dagger\dagger}$ \cite{Kim2018TextualEF} & vehicle control for self-driving cars & crowd &  $\sim$26K \\
        \textsc{\small VQA-E}$^{\dagger\dagger}$ \cite{Li2018VQAEEE} & visual QA & auto &  $\sim$270K \\
        \textsc{\small VQA-X}$^{\dagger\dagger}$ \cite{park2018multimodal} & visual QA  & crowd &  28,180 (1 or 3)\\
        \textsc{\small ACT-X}$^{\dagger\dagger}$ \cite{park2018multimodal} & activity recognition  & crowd &  18,030 (3) \\
        \citet{Ehsan2019AutomatedRG}$^{\dagger\dagger}$ & playing arcade games  & crowd &  2000 \\
        \textsc{\small VCR}$^{\dagger\dagger}$ \cite{Zellers2019FromRT} & visual commonsense reasoning  & crowd & $\sim$290K \\
        \textsc{\small E-SNLI-VE}$^{\dagger\dagger}$ \cite{do2020snli}  & visual-textual entailment  & crowd & 11,335 (3)$^{\ddagger}$ \\
        \textsc{\small ESPRIT}$^{\dagger\dagger}$ \cite{rajani-etal-2020-esprit} &  reasoning about qualitative physics  & crowd &  2441 (2) \\ 
        \textsc{\small VLEP}$^{\dagger\dagger}$ \cite{lei-etal-2020-likely} & future event prediction & auto + crowd & 28,726  \\
        \textsc{\small EMU}$^{\dagger\dagger}$ \cite{Da2020EditedMU} & reasoning about manipulated images & crowd &  48K \\
        \arrayrulecolor{black}\bottomrule
    \end{tabular}}
    \caption{Overview of \exnlp datasets with \textbf{free-text explanations} for textual and visual-textual tasks (marked with $\dagger\dagger$ and placed in the lower part). Values in parentheses indicate number of explanations collected per instance (if $>$ 1). 
    $\ddagger$ A subset of the original dataset that is annotated. $\ddagger\ddagger$ Subset publicly available. $\ast$ Authors semantically parse the collected explanations.
    }
    \label{table:free_text_overview}
\end{table*}

\section{Overview of Existing Datasets}
\label{sec:survey_datasets}


We overview currently available \exnlp datasets by explanation type: highlights (Table \ref{table:highlights_overview}), free-text explanations (Table \ref{table:free_text_overview}), and structured explanations (Table \ref{table:structured_overview}). Besides \textsc{\small SCAT} \cite{Yin2021DoCT}, to the best of our knowledge, all existing datasets are in English. The authors of $\sim$66\% papers cited in Tables \ref{table:highlights_overview}--\ref{table:structured_overview} report the dataset license in the paper or a repository, and 45.61\% use \emph{common} permissive licenses; for more information see Appendix \ref{sec:appendix_license}. See Appendix \ref{sec:appendix_collection} for collection details.

For each dataset, we report the number of instances (input-label pairs) and the number of explanations per instance (if $>$ 1).
The annotation procedure used to collect each dataset is reported as: crowd-annotated (``crowd''); automatically annotated through a web-scrape, database crawl, or merge of existing datasets (``auto''); or annotated by others (``experts'', ``students'', or ``authors''). 
Some authors perform semantic parsing on collected explanations (denoted with $*$); we classify them by the dataset type before parsing and list the collection type as ``crow + authors''. Tables \ref{table:highlights_overview}-\ref{table:structured_overview} elucidate that the dominant collection paradigm ($\geq$90\%) 
is via human (crowd, student, author, or expert) annotation.

\paragraph{Highlights (\autoref{table:highlights_overview})} The granularity of highlights depends on the task they are collected for. The majority of authors do not place a restriction on granularity, allowing 
words, phrases, or sentences of the original input document to be selected. The coarsest granularity in Table \ref{table:highlights_overview} is one or more paragraphs in a longer document \cite{kwiatkowski-etal-2019-natural, chalkidis-etal-2021-paragraph}. We exclude datasets that include an associated document as evidence without specifying the location of the explanation within the document (namely document retrieval datasets). We exclude \textsc{BeerAdvocate} \cite{mcauley2012learning} because it has been retracted. 

Some highlights are re-purposed from annotations for a different task. For example, \textsc{\small MultiRC} \cite{khashabi-etal-2018-looking} contains sentence-level highlights that indicate justifications of answers to questions. However, they were originally collected 
for the authors to
assess that each question in the dataset requires multi-sentence reasoning to answer.
Another example 
is \textsc{\small{Stanford Sentiment Treebank}} \cite[\textsc{\small{SST}};][]{socher-etal-2013-recursive} which contains crowdsourced sentiment annotations for word phrases extracted from movie reviews \cite{pang-lee-2005-seeing}.
Word phrases that have the same sentiment label as the review 
can be heuristically merged to get phrase-level highlights \cite{carton-etal-2020-evaluating}. Other highlights in Table \ref{table:highlights_overview} are collected by instructing annotators. Instead of giving these instructions verbatim, their authors typically describe them concisely, e.g., they say annotators are asked to highlight words justifying, constituting, indicating, supporting, or determining the label, or words that are essential, useful, or relevant for the label. The difference in wording of these instructions affects how people annotate explanations. In \sect{sec:collecting_highlights}, we discuss how one difference in annotation instructions (requiring comprehensiveness or not) can be important. 



\paragraph{Free-Text Explanations (\autoref{table:free_text_overview})}

This is a popular explanation type for both textual and visual-textual tasks, shown in the first and second half of the table, respectively.
Most free-text explanations
are generally no more than a few sentences per instance. One exception is \textsc{\small LIAR-PLUS} \cite{alhindi-etal-2020-fact}, which 
contains the conclusion paragraphs of web-scraped human-written fact-checking summaries. 

\paragraph{Structured Explanations (\autoref{table:structured_overview})}
\label{sec:structured}

Structured explanations take on dataset-specific forms. One common approach is to construct a chain of facts that detail the reasoning steps to reach an answer (``chains of facts''). Another is to place constraints on the textual explanations that annotators can write, such as requiring the use 
of certain variables in the input (``semi-structured text'').

\begin{table*}[t]
    \centering
    \resizebox{\textwidth}{!}{
    \begin{tabular}{llllrrr}
    \toprule
        \textbf{Dataset} & \textbf{Task} & \textbf{ Explanation Type} & \textbf{Collection} & \textbf{\# Instances}\\        \midrule
        \textsc{\small WorldTree V1} \cite{jansen-etal-2018-worldtree} & science exam QA & explanation graphs & authors & 1,680 \\ 
        \textsc{\small OpenBookQA} \cite{mihaylov-etal-2018-suit} & open-book science QA  & 1 fact from \textsc{\small WorldTree}  & crowd  & 5,957 \\
        \citet{yang-etal-2018-commonsense}$^{\dagger\dagger}$ & action recognition & lists of relations + attributes & crowd & 853 \\
        \textsc{\small WorldTree V2} \cite{xie-etal-2020-worldtree} & science exam QA & explanation graphs & experts & 5,100 \\ 
        \textsc{\small QED} \cite{lamm2020qed} & reading comp. QA & inference rules &  authors & 8,991 \\
        \textsc{\small QASC} \cite{khot2020qasc} & science exam QA  & 2-fact chain & authors + crowd & 9,980 \\
        \textsc{\small eQASC} \cite{jhamtani-clark-2020-learning} & science exam QA & 2-fact chain & auto + crowd  & 9,980 ($\sim$10) \\
        + \textsc{\small Perturbed} & science exam QA & 2-fact chain & auto + crowd  & n/a$^\ddagger$ \\
        \textsc{\small eOBQA} \cite{jhamtani-clark-2020-learning} & open-book science QA & 2-fact chain & auto + crowd & n/a$^\ddagger$ \\
        \citet{ye-etal-2020-teaching}$^\ast$ & \textsc{\small SQuAD} QA & semi-structured text & crowd + authors & 164 \\
        \citet{ye-etal-2020-teaching}$^\ast$ & \textsc{\small NaturalQuestions} QA & semi-structured text & crowd + authors & 109 \\
        \textsc{\small R$^4$C} \cite{inoue-etal-2020-r4c} &  reading comp. QA & chains of facts & crowd & 4,588 (3) \\
        \textsc{\small StrategyQA} \cite{geva2021strategy} & implicit reasoning QA & reasoning steps w/ highlights & crowd & 2,780 (3) \\
        \textsc{\small TriggerNER} & named entity recognition & groups of highlighted tokens & crowd & $\sim$7K (2) \\
        \arrayrulecolor{black}\bottomrule
    \end{tabular}}
    \caption{Overview of \exnlp datasets with \textbf{structured explanations} (\sect{sec:case_study_cose_esnli}). Values in parentheses indicate number of explanations collected per instance (if $>$ 1). $\dagger\dagger$ Visual-textual dataset. $\ast$ Authors semantically parse the collected explanations. $\ddagger$ Subset of instances annotated with explanations is not reported. Total \# of explanations is 855 for \textsc{\small eQASC Perturbed} and 998 for \textsc{\small eOBQA}.}
    \label{table:structured_overview}
\end{table*}

The \textsc{\small WorldTree} datasets \cite{jansen-etal-2018-worldtree, xie-etal-2020-worldtree} propose explaining elementary-school science questions with a combination of chains of facts and semi-structured text, termed ``explanation graphs''. The facts are individual sentences written by the authors that are centered around 
a set of shared relations 
and properties. 
Given the chain of facts for an instance (6.3 facts on average), the authors can construct an explanation graph by linking 
shared words in the question, answer, and explanation. 

\textsc{\small OpenBookQA} \cite[\textsc{\small OBQA};][]{mihaylov-etal-2018-suit} uses single \textsc{\small WorldTree} facts  to prime annotators to write QA pairs. 
Similarly, each question in \textsc{\small QASC} \cite{khot2020qasc} contains two associated science facts from a corpus selected by human annotators who wrote the question.
\citet{jhamtani-clark-2020-learning} extend \textsc{\small OBQA} and \textsc{\small QASC} with  two-fact chain explanation annotations, which are automatically extracted from a fact corpus and validated with crowdsourcing. 
The resulting datasets, \textsc{\small eQASC} and \textsc{\small eOBQA}, contain multiple valid and invalid explanations per instance, as well as perturbations for robustness testing (\textsc{\small eQASC-perturbed}).

A number of structured explanation datasets supplement datasets for reading comprehension. \citet{ye-etal-2020-teaching} collect semi-structured explanations for \textsc{\small NaturalQuestions} \cite{kwiatkowski-etal-2019-natural} and \textsc{\small SQuAD} \cite{rajpurkar-etal-2016-squad}. They require annotators to use phrases in both the input question and context, and limit them to a small set of connecting expressions. 
\citet{inoue-etal-2020-r4c} collect \textsc{\small R$^4$C}, fact chain explanations for \textsc{\small HotpotQA} \cite{yang-etal-2018-hotpotqa}. 
\citet{lamm2020qed} collect explanations for \textsc{\small NaturalQuestions} that follow a linguistically-motivated form (see the example in \autoref{tab:explanation_examples}). We discuss structured explanations further in \sect{sec:case_study_cose_esnli}.


\section{Link Between \textsc{ExNLP} Data, Modeling, and Evaluation Assumptions}
\label{sec:collecting_highlights}

All three parts of the machine learning pipeline (data collection, modeling, and evaluation) are inextricably linked. In this section, we discuss what \exnlp modeling and evaluation research reveals about the qualities of available \exnlp datasets, and how best to collect such datasets in the future.

Highlights are usually evaluated following two criteria: (i) \emph{plausibility}, according to humans, how well a highlight supports a predicted label \cite{Yang2019EvaluatingEW, deyoung-etal-2020-eraser}, 
and (ii) \emph{faithfulness} or \emph{fidelity}, how accurately a highlight represents the model's decision process \cite{AlvarezMelis2018TowardsRI, wiegreffe-pinter-2019-attention}. Human-annotated highlights (Table \ref{tab:explanation_examples_two}) are used to measure the plausiblity of model-produced highlights: the higher the overlap between the two, the more plausible model highlights are considered.  
On the other hand, a highlight that is both sufficient (implies the prediction, \sect{sec:terminology}; first example in Table \ref{tab:explanation_examples_two}) and comprehensive (its complement in the input does \emph{not} imply the prediction, \sect{sec:terminology}; second example in Table \ref{tab:explanation_examples_two}) is regarded as faithful to the prediction it explains \cite{deyoung-etal-2020-eraser, carton-etal-2020-evaluating}. 
Since human-annotated highlights are used only for evaluation of plausibility but not faithfulness, one might expect that the measurement and modeling of faithfulness cannot influence how human-authored explanations should be collected. In this  section, we show that this expectation might lead to collecting highlights that are unfitting for the goals (ii) and (iii) in \sect{sec:introduction}. 




Typical instructions for collecting highlights encourage sufficiency and compactness, but not comprehensiveness.  For example, \citet{deyoung-etal-2020-eraser} deem \textsc{\small MovieReviews} and \textsc{\small EvidenceInference} highlights non-comprehensive. \citet{carton-etal-2020-evaluating} expect that \textsc{\small FEVER} highlights are non-comprehensive, in contrast to \citet{deyoung-etal-2020-eraser}. Contrary to the characterization of both of these work,  
we observe that the \esnli authors collect non-comprehensive highlights, since they instruct annotators to highlight only words in the hypothesis (and not the premise) for neutral pairs, and consider contradiction/neutral explanations correct if at least one piece of evidence in the input is highlighted. 
Based on these discrepancies in characterization, we first conclude that post-hoc assessment of comprehensiveness from a general description of data collection is error-prone.

Alternatively, \citet{carton-etal-2020-evaluating} empirically show that available human highlights are not necessarily sufficient nor comprehensive for predictions of  
\emph{highly accurate} models. This suggests that the same might hold for gold labels, leading us to ask: are gold highlights in existing datasets flawed? %

Let us first consider insufficiency. Highlighted input elements taken together have to reasonably indicate the label. %
Otherwise, a highlight is an invalid explanation. %
Consider two datasets whose sufficiency \citet{carton-etal-2020-evaluating} found to be most concerning: 
neutral \esnli pairs and 
no-attack \textsc{\small WikiAttack} examples. Neutral \esnli cases are not justifiable by highlighting because they are obtained only as an intermediate step to collecting free-text explanations, and only free-text explanations truly justify a neutral label 
\cite{camburu2018snli}. Table \ref{tab:explanation_examples_two} shows one \esnli  highlight that is not sufficient. No-attack  \textsc{\small WikiAttack} examples are not explainable by highlighting because the absence of offensive content justifies the no-attack label, and this absence cannot be highlighted. We recommend (i) avoiding human-annotated highlights with low sufficiency when evaluating and collecting highlights, and (ii) assessing whether the true label can be explained by highlighting. 

Consider a highlight that is non-comprehensive because it is redundant with its complement in the input (e.g., a word appears multiple times, but only one occurrence is highlighted).
Highlighting only one occurrence of ``great'' is a valid justification, but quantifying faithfulness of this highlight is hard because the model might rightfully use the unhighlighted occurrence of ``great'' to make the prediction. Thus, comprehensiveness is modeled to make faithfulness evaluation feasible. 
Non-comprehensiveness of human highlights, however, hinders evaluating plausibility of comprehensive model highlights since model and human highlights do not match by design. To be able to evaluate both plausibility and faithfulness, we should annotate comprehensive human highlights. We summarize these observations in Figure \ref{fig:reconcile} in Appendix \ref{appendix:additional_info}.



Mutual influence of data and modeling assumptions also affects free-text explanations. %
For example, the \esnli guidelines 
have far more 
constraints than the \cose guidelines, 
such as requiring self-contained explanations. %
\citet{Wiegreffe2020MeasuringAB} show that such data collection decisions can influence modeling assumptions. 
This is not an issue per se, but we should be cautious that \exnlp data collection decisions do not popularize explanation properties as \emph{universally necessary} when they are not, e.g., that free-text explanations should be understandable without the original input or that highlights should be comprehensive. %
We believe this could be avoided with better documentation, e.g., with additions to a standard datasheet \cite{Gebru2018DatasheetsFD}. Explainability fact sheets have been proposed for models \cite{sokol2020explainability}, but not for datasets. 
For example, an \esnli datasheet could note that self-contained explanations were required during data collection, but that this is not a necessary property of a valid free-text explanation. 
A dataset with comprehensive highlights should emphasize 
that comprehensiveness is required to simplify faithfulness evaluation.

\textbf{Takeaways} 

\begin{compactenum}
\item  It is important to precisely report how explanations were collected, e.g., by giving access to the annotation interface, screenshotting it, or giving the annotation instructions verbatim.
\item Sufficiency is necessary for highlights, and \exnlp researchers should avoid human-annotated highlights with low sufficiency for evaluating and developing highlights.
\item Comprehensiveness isn't necessary for a valid highlight, it is a means to quantify faithfulness.
\item Non-comprehensive human-annotated highlights cannot be used to automatically evaluate plausibility of highlights that are constrained to be comprehensive. In this case, \exnlp researchers should collect and use comprehensive human-annotated highlights.
\item 
Researchers should not make (error-prone) post-hoc estimates of comprehensiveness of human-annotated highlights from datasets' general descriptions. 
\item \exnlp researchers should be careful to not popularize their data collection decisions as universally necessary. We advocate for documenting all constraints on collected explanations in a datasheet, highlighting whether each constraint is necessary for explanation to be valid or not, and noting how each constraint might affect modeling and evaluation. 
\end{compactenum}

\section{Rise of Structured Explanations}
\label{sec:case_study_cose_esnli}

The merit of free-text explanations is their expressivity, which can come at the costs of underspecification and inconsistency due to the difficulty of quality control (stressed by the creators of two popular free-text explanation datasets: 
\esnli and \cose). In this section, we highlight and challenge one prior approach to overcoming these difficulties: discarding template-like free-text explanations. 

We gather crowdsourcing guidelines for the above-mentioned datasets in Tables \ref{tab:instructions_esnli}--\ref{tab:instructions_cose} in Appendix and compare them. %
We observe two notable similarities between the guidelines for the above-mentioned datasets. %
First, both asked annotators to first highlight input words and then formulate a free-text explanation from them, to control quality. %
Second, template-like explanations are discarded because they are deemed uninformative. %
The \esnli authors assembled a list of 56 templates (e.g., ``There is $\langle$hypothesis$\rangle$'') to identify explanations whose edit distance to one of the templates is $<$10. %
They re-annotate the detected template-like explanations (11\% in the entire dataset). %
The \cose authors discard sentences ``$\langle$answer$\rangle$ is the only option that is correct/obvious'' (the only given example of a template). %
Template explanations concern researchers because they can result in artifact-like behaviors in certain modeling architectures.
For example, a model which predicts a task output from a generated explanation can produce explanations that are plausible to a human user and give the impression of making label predictions on the basis of this explanation. %
However, it is possible that the model learns to ignore the semantics of the explanation and instead makes predictions based on the explanation's template type \cite{kumar-talukdar-2020-nile, jacovi2020aligning}. %
In this case, the semantic interpretation of the explanation (that of a human reader) is not faithful (an accurate representation of the model's decision process).

Despite re-annotating, \citet{camburu-etal-2020-make} report that \esnli explanations still largely follow 28 label-specific templates (e.g., an entailment template ``X is another form of Y'') even after re-annotation. Similarly, \citet{Brahman2020LearningTR} report that models trained on gold \esnli explanations generate template-like explanations for the  defeasible NLI task. 
These findings lead us to ask: what are the differences between templates considered uninformative and filtered out,  and those identified by \citet{camburu-etal-2020-make, Brahman2020LearningTR} that remain after filtering? Are \emph{all} template-like explanations uninformative?

Although prior work 
indicates that template-like explanations are undesirable, most recently, structured explanations have been intentionally collected (see Table \ref{table:structured_overview}; \sect{sec:structured}). 
What these studies share is that they acknowledge structure as \emph{inherent} to explaining the tasks they investigate. Related work \cite[\glucose;][]{mostafazadeh-etal-2020-glucose} takes the matter further, arguing that explanations should not be entirely free-form. 
Following \glucose, we recommend running pilot studies to explore how people define and generate explanations for a task \emph{before} collecting free-text explanations for it. 
If they reveal that informative human explanations are naturally structured, incorporating the structure in the annotation scheme is useful since the structure is natural to explaining the task. %
This turned out to be the case with NLI; \citet{camburu-etal-2020-make} report: ``Explanations in \esnli largely follow a set of label-specific templates. This is a \emph{natural consequence of the task} and 
dataset''.
We recommend embracing the structure when possible, but also encourage creators of datasets with template-like explanations to highlight in a dataset datasheet (\sect{sec:collecting_highlights}) that template structure can influence downstream modeling decisions. There is no all-encompassing definition of explanation, and researchers could consult domain experts or follow literature from other fields to define an appropriate explanation in a task-specific manner, such as in \glucose \cite{mostafazadeh-etal-2020-glucose}. For conceptualization of explanations in different fields see \citet{Tiddi2015AnOD}.



Finally, what if pilot studies do not reveal any obvious structure to human explanations of a task? Then we need to do our best to control the quality of free-text explanations because low dataset quality is a bottleneck to building high-quality models. \cose is collected with notably less annotation constraints and quality controls than \esnli, and has annotation issues that some have deemed make the dataset unusable \cite{narang2020wt5}; see examples in Table \ref{tab:instructions_cose} of \autoref{appendix:additional_info}. As exemplars of quality control, we point the reader to the annotation guidelines of \textsc{\small VCR} \cite{Zellers2019FromRT} in Table \ref{tab:overview_vcr} 
and \glucose \cite{glucoseCollection}. In \sect{sec:quality} and \sect{sec:related_quality}, we give further task-agnostic recommendations for collecting high-quality \exnlp datasets, applicable to all three explanation types. 

\textbf{Takeaways} 

\begin{compactenum}
\item \exnlp researchers should study how people define and generate explanations for the task before collecting free-text explanations. 
\item If pilot studies show that explanations are naturally structured, embrace the structure. 
\item There is no all-encompassing definition of explanation. Thus, \exnlp researchers could consult domain experts or follow literature from other fields to define an appropriate explanation form, and these matters should be open for debate on a given task. 
\end{compactenum}

\section{Increasing Explanation Quality}
\label{sec:quality}

When asked to write free-text sentences from scratch for a table-to-text annotation task outside \exnlp, 
\citet{parikh-etal-2020-totto} note that crowdworkers produce ``vanilla targets that lack [linguistic] variety''. Lack of variety can result in annotation artifacts, which are prevalent in the popular \snli \cite{bowman-etal-2015-large} and \mnli \cite{williams-etal-2018-broad} datasets \cite{poliak-etal-2018-hypothesis,  gururangan-etal-2018-annotation, tsuchiya-2018-performance}, among others
\cite{geva-etal-2019-modeling}. These authors demonstrate the harms of such artifacts: models can overfit to them, leading to both performance over-estimation and problematic generalization behaviors.

Artifacts can occur from poor-quality annotations and inattentive annotators, both of which have been on the rise on crowdsourcing platforms \cite{chmielewski2020mturk, arechar2020turking, narang2020wt5}. 
To mitigate artifacts, both increased \textbf{diversity of annotators} and \textbf{quality control} are needed. We focus on quality control here and diversity in \sect{sec:related_quality}.

\subsection{A Two-Stage Collect-And-Edit Approach}
\label{sec:two_stage_collection}

While ad-hoc methods can improve quality \cite{camburu2018snli, Zellers2019FromRT, glucoseCollection}, an effective and generalizable method is to collect annotations in two stages. 
A two-stage methodology has been applied by a small minority of \exnlp dataset papers \cite{jhamtani-clark-2020-learning, zhang-etal-2020-winowhy, Zellers2019FromRT}, who first compile explanation candidates automatically or from crowdworkers, and secondly perform quality-control by having other crowdworkers 
assess the quality of the collected explanations  (we term this \CollectAndJudge). %
Judging improves the overall quality of the final dataset by removing low-quality instances, and additionally allows authors to release quality ratings for each instance.

Outside \exnlp, \citet{parikh-etal-2020-totto} use an extended version of this approach (that we term \CollectAndEdit): they generate a noisy automatically-extracted dataset for the table-to-text generation task, and then ask annotators to edit the datapoints. %
\citet{bowman-etal-2020-new} use this approach to re-collect NLI hypotheses, and find, crucially, that having annotators edit rather than create hypotheses reduces artifacts in a subset of \mnli. %
In XAI, \citet{kutlu2020annotator} collect highlight explanations for Web page ranking with annotator editing. %
We advocate expanding the \CollectAndJudge approach for explanation collection to \CollectAndEdit. %
This has potential to increase linguistic diversity via multiple annotators per-instance, reduce individual annotator biases, and perform quality control. Through a case study of two multimodal free-text explanation datasets, we will demonstrate that collecting explanations automatically without human editing (or at least judging) can lead to artifacts.


\esnlivte \cite{do2020snli} and \vqae \cite{Li2018VQAEEE} are two visual-textual datasets for entailment and question-answering, respectively.
\esnlivte combines annotations of two datasets: (i) \snlive \cite{xie2019visual}, collected by replacing the textual premises of \snli \cite{bowman-etal-2015-large} with \flickr images \cite{young-etal-2014-image}, and (ii) \esnli \cite{camburu2018snli}, a dataset of crowdsourced explanations for \snli. This procedure is possible because 
every \snli premise was originally the caption of a \flickr photo. However, since \snli's hypotheses were collected from crowdworkers who did not see the original images,
the photo replacement process results in a significant number of errors 
\cite{vu-etal-2018-grounded}. \citet{do2020snli} re-annotate labels and explanations for the neutral pairs in the validation and test sets of \snlive. However, it has been argued that the dataset remains low-quality for training models 
due to artifacts in the entailment and the neutral class' training sets \cite{marasovic-etal-2020-natural}. With a full \edit approach, we expect that these artifacts would be significantly reduced, and the resulting dataset could have quality on-par with \esnli. 
Similarly, the \vqae dataset \cite{Li2018VQAEEE} 
converts image captions from the \textsc{\small{VQA v2.0}} dataset \cite{Goyal2017MakingTV} into explanations,
but a notably lower plausibility compared to a carefully-crowdsourced \textsc{\small VCR} explanations is reported in \cite{marasovic-etal-2020-natural}. 

Both \esnlivte and \vqae present novel and cost-effective ways to produce large \exnlp datasets for new tasks, but also show the quality tradeoffs of automatic collection. Strategies such as crowdsourced judging and editing, even on a small subset, can reveal and mitigate such issues.

\subsection{Teach and Test the Underlying Task}
\label{sec:teaching_task}

In order to both create and judge explanations, annotators must understand the underlying task and label-set well. In most cases, this necessitates teaching and testing 
the task.
Prior work outside of \exnlp has noted the difficulty of scaling annotation to crowdworkers for complex linguistic tasks \cite{roit-etal-2020-controlled, elazar-etal-2020-extraordinary, pyatkin-etal-2020-qadiscourse, mostafazadeh-etal-2020-glucose}.
To increase annotation quality, 
these works provide intensive training to crowdworkers, including personal feedback. 
Since label understanding is a prerequisite for explanation collection, task designers should consider relatively inexpensive strategies such as qualification tasks and checker questions. 
This need is correlated with the difficulty and domain-specificity of the task, as elaborated above. 

Similarly, people cannot explain all tasks equally well and even after intensive training they might struggle to explain tasks such as deception detection and recidivism prediction \cite{Nisbett1977TellingMT}. Human explanations for such tasks might be limited in serving the three goals outlined in \sect{sec:introduction}.

\subsection{Addressing Ambiguity}
\label{subsec:unsure}


Data collectors often collect explanations post-hoc, i.e., annotators are asked to explain labels assigned by a system or other annotators. The underlying assumption is that the explainer believes the assigned label to be correct or at least likely (there is no task ambiguity). However, this assumption has been shown to be inaccurate (among others) for 
relation extraction \cite{aroyo2013crowd}, natural language inference \cite{pavlick-kwiatkowski-2019-inherent, nie-etal-2020-learn},  and complement coercion \cite{elazar-etal-2020-extraordinary}, and the extent to which it is true likely varies by task, instance, and annotator. If an annotator is uncertain about a label, their explanation may be at best a hypothesis and at worst a guess. HCI research encourages leaving room for
ambiguity rather than forcing  raters into binary decisions, which can result in poor or inaccurate labels \cite{sambasivantalk}. 

To ensure explanations reflect human decisions as closely as possible, it is ideal to collect both labels and explanations from the same annotators. Given that this is not always possible, including a checker question to assess whether an explanation annotator agrees with a label is a good alternative.

\textbf{Takeaways}

\begin{compactenum}
\item  Using a \CollectAndEdit method can reduce individual annotator biases, perform quality control, and potentially reduce dataset artifacts.
\item Teaching and testing the underlying task and addressing ambiguity can improve data quality.
\end{compactenum}

\section{Increasing Explanation Diversity}
\label{sec:related_quality}

Beyond quality control, increasing annotation diversity is another task-agnostic means to mitigate artifacts and collect more representative data. We elaborate on suggestions from related work (inside and outside \exnlp) here.

\subsection{Use a Large Set of Annotators}
\label{sec:num_annotators}

Collecting representative data entails ensuring that a handful of annotators do not dominate data collection.
Outside \exnlp, \citet{geva-etal-2019-modeling} report that recruiting only a small pool of annotators (1 annotator per 100--1000 examples) allows models 
to overfit on annotator characteristics. Such small annotator pools exist in \exnlp---for instance, \esnli reports an average of 860 explanations written per worker. The occurrence of the incorrect explanation ``rivers flow trough valleys'' for 529 different instances in \cose v1.11 is likely attributed to a single annotator. \citet{al-kuwatly-etal-2020-identifying} find that demographic attributes can predict annotation differences. Similarly, \citet{davidson-etal-2019-racial, sap-etal-2019-risk} show that annotators often consider African-American English writing to be disproportionately offensive.\footnote{In another related study, 82\% of annotators reported their race as white \cite{sap-etal-2020-social}. This is a likely explanation for the disproportionate annotation.} A lack of annotator representation concerns \exnlp for three reasons: explanations depend on socio-cultural background \cite{kopeckaexplainable}, annotator traits should not be predictable \cite{geva-etal-2019-modeling}, and the subjectivity of explaining leaves room for social bias to emerge.
On most platforms, annotators are not restricted to a specific number of instances. 
Verifying that no worker has annotated an excessively large portion of the dataset in addition to strategies from \citet{geva-etal-2019-modeling} can help mitigate annotator bias.
More elaborate methods for increasing annotator diversity include collecting demographic attributes or modeling annotators as a graph \cite{al-kuwatly-etal-2020-identifying, wich-etal-2020-investigating}.


\subsection{Multiple Annotations Per Instance}
\label{sec:num_annotations}

HCI research has long considered the ideal of crowdsourcing a single ground-truth as a ``myth'' that fails to account for the diversity of human thought and experience \cite{aroyo2015truth}. Similarly, \exnlp researchers should not assume there is always one correct explanation. Many of the assessments crowdworkers are asked to make when writing explanations are subjective in nature, and there are many different models of explanation based on a user's cognitive biases, social expectations, and socio-cultural background \cite{miller2019explanation}. \citet{Prasad2020ToWE} present a theoretical argument to illustrate that there are multiple ways to highlight input words to  explain an annotated sentiment label. 
\citet{camburu2018snli} find a low inter-annotator BLEU score \cite{papineni-etal-2002-bleu} between free-text explanations collected for \esnli test instances. 

If a dataset contains only one explanation when multiple are plausible, a plausible model explanation can be penalized unfairly for not agreeing with it. %
We expect that modeling multiple explanations can also be a useful learning signal. %
Some existing datasets contain multiple explanations per instance (last column of Tables \ref{table:highlights_overview}--\ref{table:structured_overview}). Future \exnlp data collections should do the same if there is subjectivity in the task or diversity of correct explanations (which can be measured via inter-annotator agreement). If annotators exhibit low agreement between explanations deemed as plausible, this can reveal a diversity of correct explanations for the task, which should be considered in modeling and evaluation.



\subsection{Get Ahead: Add Contrastive and Negative Explanations}
\label{sec:coverage}


The machine learning community has championed modeling 
\emph{contrastive explanations} that justify why a prediction was made 
\emph{instead of} another, 
to align more closely with human explanation \cite{dhurandhar2018explanations, hendricks2018generating, miller2019explanation}. 
Most recently, methods have been proposed in NLP to produce contrastive edits of the input as explanations \cite{ross2020explaining, Yang2020GeneratingPC, Wu2021PolyjuiceAG, jacovi2020aligning}. 
Outside of \exnlp, datasets with contrastive edits have been collected to assess and improve robustness of NLP models \cite{kaushik2019learning, gardner-etal-2020-evaluating, li-etal-2020-linguistically} and might be used for explainability too.

Just as highlights are not sufficiently intelligible for complex tasks, the same might hold for contrastive input edits. 
To the best of our knowledge, there is no dataset that contains contrastive free-text or structured explanations. 
These could take the form of (i) collecting explanations that answer the question ``why...instead of...'', or (ii) collecting explanations for other labels besides the gold label, to be used as an additional training signal. 
A related annotation paradigm is to collect \emph{negative  explanations}, i.e., explanations that are invalid for an (input, gold label) pair. Such examples can improve \exnlp models by providing supervision of what is \emph{not} a correct explanation \cite{schuff-etal-2020-f1}. A human \judge or \edit phase automatically gives negative explanations: the low-scoring instances (former) or instances pre-editing (latter) \cite{, jhamtani-clark-2020-learning, zhang-etal-2020-winowhy}. 

\textbf{Takeaways}

\begin{compactenum}
\item To increase annotation diversity, a large set of annotators, multiple annotations per instance, and collecting explanations that are most useful to the needs of end-users are important.
\item Reporting inter-annotator agreement with plausibility of annotated explanations is useful to known whether there is a natural diversity of explanations for the task and should the diversity be considered for modeling and evaluation.
\end{compactenum}

\section{Conclusions} 

We have presented a review of existing datasets for \exnlp research, highlighted discrepancies in data collection that can have downstream modeling effects, and synthesized the literature both inside and outside \exnlp into a set of recommendations for future data collection. 

We note that a majority of the work reviewed in this paper has originated in the last 1-2 years, indicating an explosion of interest in collecting datasets for \exnlp{}. We provide reflections for current and future data collectors in an effort to promote standardization and consistency. This paper also serves as a starting resource for newcomers to \exnlp{}, and, we hope, a starting point for further discussions. 

\section*{Acknowledgements}
We are grateful to Yejin Choi, Peter Clark, Gabriel Ilharco, Alon Jacovi, Daniel Khashabi, Mark Riedl, Alexis Ross, and Noah Smith for valuable feedback.

\bibliographystyle{plainnat}
\bibliography{neurips_2021,anthology}

\appendix
\clearpage

\section{Complementing Information}
\label{appendix:additional_info}
We provide the following additional illustrations and information that
complement discussions in the main paper: 
\begin{itemize}
\item Details of dataset licenses in Appendix \ref{sec:appendix_license}.
\item Details of dataset collection in Appendix \ref{sec:appendix_collection}.
\item An illustration of connections between assumptions made in the development of self-explanatory highlighting models (discussed in \sect{sec:collecting_highlights}) is shown in Figure \ref{fig:reconcile}. 
\item Overviews of quality measures and outcomes in \esnli, \cose, and \textsc{\small VCR} in Tables \ref{tab:instructions_esnli}--\ref{tab:overview_vcr}. 
\item A discussion of explanation and commonsense reasoning in \autoref{sec:definition}.
\end{itemize}

\section{Dataset Licenses} 
\label{sec:appendix_license}

The authors of 33.96\% papers cited in Tables \ref{table:highlights_overview}--\ref{table:structured_overview} do \textbf{not} report the dataset license in the paper or a repository;  45.61\% use \emph{common} permissive licenses such as Apache 2.0, MIT, CC BY-SA 4.0, CC BY-SA 3.0, BSD 3-Clause ``New'' or ``Revised'' License, BSD 2-Clause ``Simplified'' License, CC BY-NC 2.0, CC BY-NC-SA, GFDL, and CC0 1.0 Universal. We overview the rest:
\begin{itemize}
    \item \textsc{WikiQA}: ``Microsoft Research Data License Agreement for Microsoft Research WikiQA Corpus''
    \item \textsc{MultiRC}: ``Research and Academic Use License''
    \item \citet{hanselowski-etal-2019-richly}: A data archive is under \textbf{Copyright}.
    \item \textsc{CoQA}: ``Children's stories are collected from MCTest \cite{richardson-etal-2013-mctest} which comes with MSR-LA license. Middle/High school exam passages are collected from RACE \cite{lai-etal-2017-race} which comes with its own license.'' The rest of the dataset is under permissive licenses: BY-SA 4.0 and Apache 2.0. 
    \item \citet{wang2019learning}: The part of the dataset that is built on on \textsc{TACRED} \cite{zhang-etal-2017-position} cannot be distributed (under ``LDC User Agreement for Non-Members'') and the license for the rest of dataset is not specified.
    \item \textsc{BDD-X}: ``UC Berkeley's Standard Copyright and Disclaimer Notice''
    \item \textsc{VCR}: ``Dataset License Agreement''
    \item \textsc{VLEP}: ``VLEP Dataset Download Agreement''
    \item \textsc{WorldTree V1}: ``End User License Agreement''
    \item \textsc{WorldTree V2}: ``End User License Agreement''
    \item \textsc{ECQA}: ``Community Data License Agreement - Sharing - Version 1.0''
\end{itemize}

\section{Dataset Collection}
\label{sec:appendix_collection}

To collect the datasets, we used our domain expertise, having previously published work using highlights and free-text explanations, to construct a seed list of datasets. In the year prior to submission, we augmented this list as we encountered new publications and preprints. We then searched the ACL Anthology (\url{https://aclanthology.org}) for the terms ``explain'', ``interpret'', ``explanation'', and ``rationale'', focusing particularly on proceedings from 2020 and onward, as the subfield has grown in popularity significantly in this timeframe. We additionally first made live the website open to public contributions 3.5 months prior to submission, and integrated all dataset suggestions we received into the tables.

\section{Explanation and Commonsense Reasoning}
\label{sec:definition}

\begin{figure}[t]
    \centering
    \includegraphics[width=0.9\textwidth]{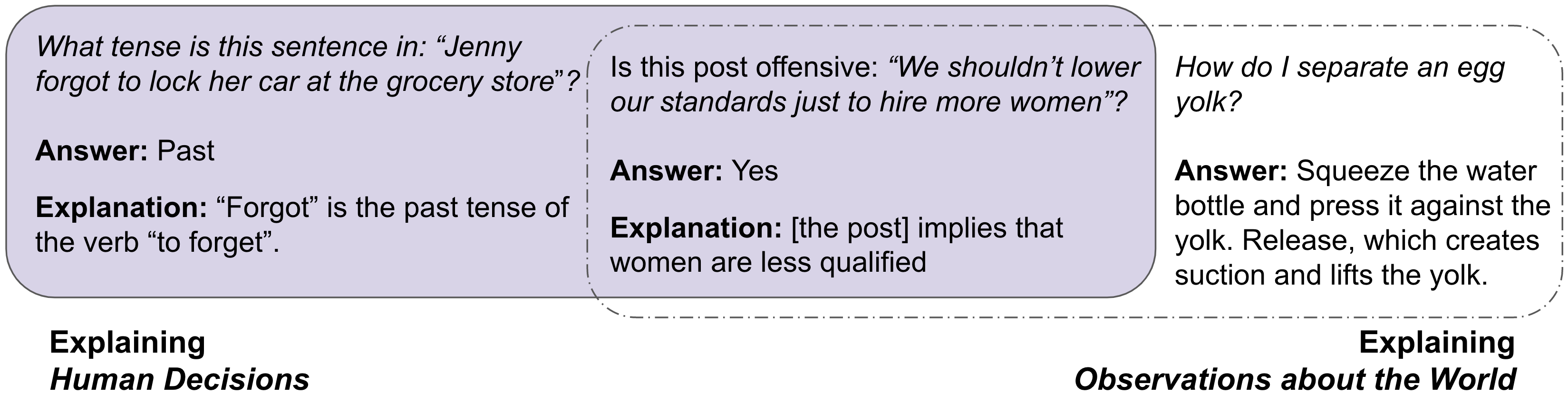}
    \caption{
    Two classes of \exnlp datasets (\sect{sec:definition}). 
    The shaded area is our 
    scope.} 
    \label{fig:definitions}
\end{figure}

The scope of our survey focuses on textual explanations that explain \emph{human decisions} (defined in the survey as task labels). There has recently emerged a set of datasets at the intersection of commonsense reasoning and explanation (such as \glucose \cite{mostafazadeh-etal-2020-glucose}). We class these datasets as explaining \emph{observed events or phenomena} in the world, where the distinction between class label and explanation is not defined. For an illustration of the difference between these datasets and those surveyed in the main paper, see \autoref{fig:definitions}.


Unlike the datasets surveyed in the paper, datasets that explain \emph{observed events or phenomena} in the world (often in the form of commonsense inferences) do not fit the three main goals of \exnlp because they do not lend themselves to task-based explanation modeling. These datasets generally do not use the term ``explanation'' \cite[\emph{inter alia}]{huang-etal-2019-cosmos, fan-etal-2019-eli5, forbes-etal-2020-social}, 
with two exceptions: \textsc{\small ART} \cite{bhagavatula2019abductive} and \textsc{\small GLUCOSE} \cite{mostafazadeh-etal-2020-glucose}. 
They produce tuples of the form (input, label), where the input is an event or observation and the label can possibly be seen as an explanation, rather than (input, label, explanation). 

Some datasets surveyed in the paper fit both categories. For instance, \textsc{\small SBIC} \cite{sap-etal-2019-risk} contains both human-annotated ``offensiveness''  labels and justifications 
of why social media posts might be considered offensive (middle of Fig.\  \ref{fig:definitions}). 
Other examples 
include predicting future events in videos \cite[\textsc{\small VLEP};][]{lei-etal-2020-likely} and answering commonsense questions about images \cite[\textsc{\small VCR};][]{Zellers2019FromRT}. Both collect observations about a real-world setting as task labels as well as explanations. We include them in our survey. 

\paragraph{A side-note on the scope.} We discuss some necessary properties of human-authored explanations (e.g., sufficiency in \sect{sec:collecting_highlights}) and conditions under which they are necessary (e.g., comprehensiveness if we wish to evaluate plausibility of model highlights that are constrained to be comprehensive; \sect{sec:collecting_highlights}), as well as properties that are previously typically considered as unwanted but we illustrate they are not necessarily inappropriate (e.g., template-like explanations in \sect{sec:case_study_cose_esnli}). However, there might be other relevant properties of human-annotated explanations that we did not discuss since we focus on discussing topics most relevant to the latest ExNLP and NLP research such as sufficiency, comprehensivness, plausibility, faithfulness, template-like explanations, and data artifacts. Moreover, as we highlight in \sect{sec:case_study_cose_esnli}, there is no all-encompassing definition of explanation and thus there  we do not expect that there is universal criteria for an appropriate explanation.

\clearpage

\begin{table*}[!h]
\resizebox{\textwidth}{!}{
\begin{tabular}{p{1.11\textwidth}}
\toprule
\makecell[c]{\textsc{Explaining Natural Language Inference (e-SNLI;}\citealt{camburu2018snli})}\\
\midrule
\makecell[l]{\textbf{General Constraints for Quality Control}}\\
Guided annotation procedure:                    \\
$\bullet$ Step 1: Annotators had to highlight words from the premise/hypothesis that are essential for the given relation.\\
$\bullet$ Step 2: Annotators had to formulate a free-text explanation using the highlighted words.\\
$\bullet$ To avoid ungrammatical sentences, only half of the highlighted words had to be used with the same spelling. \\
\makecell[l]{$\bullet$ The authors checked that the annotators also used non-highlighted words; correct explanations needs articulate a\\ \hspace{2mm} link between the keywords.}     \\
$\bullet$ Annotators had to give self-contained explanations: sentences that make sense without the premise/hypothesis.\\
$\bullet$ Annotators had to focus on the premise parts that are \emph{not} repeated in the hypothesis (non-obvious elements).\\
$\bullet$ In-browser check that each explanation contains at least three tokens.\\
$\bullet$ In-browser check that an explanation is not a copy of the premise or hypothesis.\\
\arrayrulecolor{black!20}\midrule
\makecell[l]{\textbf{Label-Specific Constraints for Quality Control}}\\ 
$\bullet$ For entailment, justifications of all the parts of the hypothesis that do not appear in the premise were required. \\
\makecell[l]{$\bullet$ For neutral and contradictory pairs, while annotators were encouraged to state all the elements that contribute to\\ \hspace{2mm} the relation, an explanation was considered correct if at least one element is stated.} \\
$\bullet$ For entailment pairs, annotators had to highlight at least one word in the premise.                 \\
$\bullet$ For contradiction pairs, annotators had to highlight at least one word in both the premise and the hypothesis.\\
\makecell[l]{$\bullet$ For neutral pairs, annotators were allowed to highlight only words in the hypothesis, to strongly emphasize the\\ \hspace{2mm} asymmetry in this relation and to prevent workers from confusing the premise with the hypothesis.}\\
\arrayrulecolor{black!20}\midrule
\makecell[l]{\textbf{Quality Analysis and Refinement}}\\            
\makecell[l]{$\bullet$ The authors graded correctness of 1000 random examples between 0 (incorrect) and 1 (correct), giving partial\\ \hspace{2mm} scores of k/n if only k out of n required arguments were mentioned.}                   \\
\makecell[l]{$\bullet$ An explanation was rated as incorrect if it was template-like. The authors assembled a list of 56 templates that\\ \hspace{2mm} they used for identifying explanations (in the entire dataset) whose edit distance to one of the templates was $<$10.\\ \hspace{2mm} They re-annotated the detected template-like explanations (11\% in total).} \\
 \arrayrulecolor{black!20}\midrule
\makecell[l]{\textbf{Post-Hoc Observations}}\\                            
$\bullet$ Total error rate of 9.62\%: 19.55\% on entailment, 7.26\% on neutral, and 9.38\% on contradiction.    \\
$\bullet$ In the large majority of the cases, that authors report it is easy to infer label from an explanation.\\
\makecell[l]{$\bullet$ \citet{camburu-etal-2020-make}: ``Explanations in e-SNLI largely follow a set of label-specific templates. This is a natural\\ \hspace{2mm} consequence of the task and the SNLI dataset and not a requirement in the collection of the e-SNLI. [...] For each\\ \hspace{2mm} label, we created a list of the most used templates that we manually identified among e-SNLI.'' They collected 28\\ \hspace{2mm} such templates.}\\
\arrayrulecolor{black}\bottomrule
\end{tabular}
}
\caption{Overview of quality control measures and outcomes in \textsc{E-SNLI}.}
\label{tab:instructions_esnli}
\end{table*}

\begin{table*}[t]
\resizebox{\textwidth}{!}{
\begin{tabular}{p{1.1\textwidth}}
\toprule
\makecell[c]{\textsc{Explaining Commonsense QA (CoS-E;} \citealt{rajani-etal-2019-explain})}\\  
\midrule
\makecell[l]{\textbf{General Constraints for Quality Control}}\\
Guided annotation procedure:  \\
$\bullet$ Step 1: Annotators had to highlight relevant words in the question that justifies the correct answer.\\
\makecell[l]{$\bullet$ Step 2: Annotators had to provide a brief open-ended explanation based on the highlighted justification that\\ \hspace{2mm} could serve as the commonsense reasoning behind the question.} \\
$\bullet$ In-browser check that annotators highlighted at least one relevant word in the question.\\
$\bullet$ In-browser check that an explanation contains at least four words.            \\
\makecell[l]{$\bullet$ In-browser check that an explanation is not a substring of the question or the answer choices without any other\\ \hspace{2mm} extra words.  }                              \\
\arrayrulecolor{black!20}\midrule
\makecell[l]{\textbf{Label-Specific Constraints for Quality Control}}\\ 
(none) \\
\arrayrulecolor{black!20}\midrule
\makecell[l]{\textbf{Quality Analysis and Refinement}}\\   
$\bullet$ The authors did unspecified post-collection checks to catch examples that are not caught by their previous filters.\\
\makecell[l]{$\bullet$ The authors removed template-like explanations, i.e., sentences ``$\langle$answer$\rangle$ is the only option that is correct \\ \hspace{2mm} obvious'' (the only provided example of a template).}     \\
 \arrayrulecolor{black!20}\midrule
\makecell[l]{\textbf{Post-Hoc Observations}}\\ 
$\bullet$ 58\% explanations (v1.0) contain the ground truth answer.\\
\makecell[l]{$\bullet$ The authors report that many explanations remain noisy after quality-control checks, but that they find them to\\ \hspace{2mm} be of sufficient quality for the purposes of their work.}   \\
\makecell[l]{$\bullet$ \citet{narang2020wt5} on v1.11: ``Many of the ground-truth explanations for CoS-E are low quality and/or \\ \hspace{2mm} nonsensical  (e.g., the question ``Little sarah didn’t think that anyone should be kissing boys. She thought that \\ \hspace{2mm} boys had what?'' with answer ``cooties'' was annotated with the explanation ``american horror comedy film \\ \hspace{2mm} directed''; or  the question ``What do you fill with ink to print?'' with answer ``printer'' was annotated with the\\ \hspace{2mm} explanation ``health complications'', etc.)''}\\
\makecell[l]{$\bullet$ Further errors exist (v1.11): The answer ``rivers flow trough valleys'' appears 529 times, and ``health complications'' \\ \hspace{2mm} 134 times, signifying copy-paste behavior by some annotators. Uninformative answers such as ``this word is the \\ \hspace{2mm} most relevant'' (and variants) appear 522 times.}\\
\arrayrulecolor{black}\bottomrule
\end{tabular}
}
\caption{Overview of quality control measures and outcomes in \textsc{CoS-E}.}
\label{tab:instructions_cose}
\end{table*}
\begin{table*}[t]
\resizebox{\textwidth}{!}{
\begin{tabular}{p{1.1\textwidth}}
\toprule
\makecell[c]{\textsc{Explaining Visual Commonsense Reasoning (VCR;} \citealt{Zellers2019FromRT})}\\  
\midrule
\makecell[l]{\textbf{General Constraints for Quality Control}}\\
$\bullet$ The authors automatically rate instance ``interestingness'' and collect annotations for the most ``interesting'' instances. \\
Multi-stage annotation procedure:  \\
$\bullet$ Step 1: Annotators had to write 1-3 questions based on a provided image (at least 4 words each).\\
$\bullet$ Step 2: Annotators had to answer each question (at least 3 words each). \\
$\bullet$ Step 3: Annotators had to provide a rationale for each answer (at least 5 words each). \\
$\bullet$ Annotators had to pass a qualifying exam where they answered some multiple-choice questions and wrote a question, answer, and rationale for a single image. The written responses were verified by the authors. \\
$\bullet$ Authors provided annotators with high-quality question, answer, and rationale examples.                             \\
$\bullet$ In-browser check that annotators explicitly referred to at least one object detected in the image, on average, in the question, answer, or rationale.\\
$\bullet$ Other in-browser checks related to the question and answer quality.\\
$\bullet$ Every 48 hours, the lead author reviewed work and provided aggregate feedback to make sure the annotators were proving good-quality responses and ``structuring rationales in the right way''. It is unclear, but assumed, that poor annotators were dropped during these checks. \\
\arrayrulecolor{black!20}\midrule
\makecell[l]{\textbf{Label-Specific Constraints for Quality Control}}\\ 
(none)\\
\arrayrulecolor{black!20}\midrule
\makecell[l]{\textbf{Quality Analysis and Refinement}}\\   
$\bullet$ The authors used a second phase to further refine some HITs. A small group of workers who had done well on the main task were selected to rate a subset of HITs (about 1 in 50), and this process was used to remove annotators with low ratings from the main task. \\
 \arrayrulecolor{black!20}\midrule
\makecell[l]{\textbf{Post-Hoc Observations}}\\ 
$\bullet$ The authors report that humans achieve over 90\% accuracy on the multiple-choice rationalization task derived from the dataset. They also report high agreement between the 5 annotators for each instance. These can be indicative of high dataset quality and low noise.\\
$\bullet$ The authors report high diversity---almost every rationale is unique, and the instances cover a range of commonsense categories.\\
$\bullet$ The rationales are long, averaging 16 words in length, another sign of quality. \\ 
$\bullet$ External validation of quality: \citet{marasovic-etal-2020-natural} find that the dataset's explanations are highly plausible with respect to both the image and associated question/answer pairs; they also rarely describe events or objects not present in the image. \\
\arrayrulecolor{black}\bottomrule
\end{tabular}
}
\caption{Overview of quality control measures and outcomes for (the rationale-collection portion) of \textsc{VCR}. The dataset instances (questions and answers) and their rationales were collected simultaneously; we do not include quality controls placed specifically on the question or answer.}
\label{tab:overview_vcr}
\end{table*}

\begin{figure}[t]
    \centering
    \subfloat[Supervised models' development. When we use human highlights as supervision, we assume that they are the gold-truth and that model highlights should match. Thus, comparing human and model highlights for plausibility evaluation is sound. However, with this basic approach we do not introduce any data or modeling properties that help faithfulness evaluation, and that remains a challenge in this setting.]{\includegraphics[width=0.7\columnwidth]{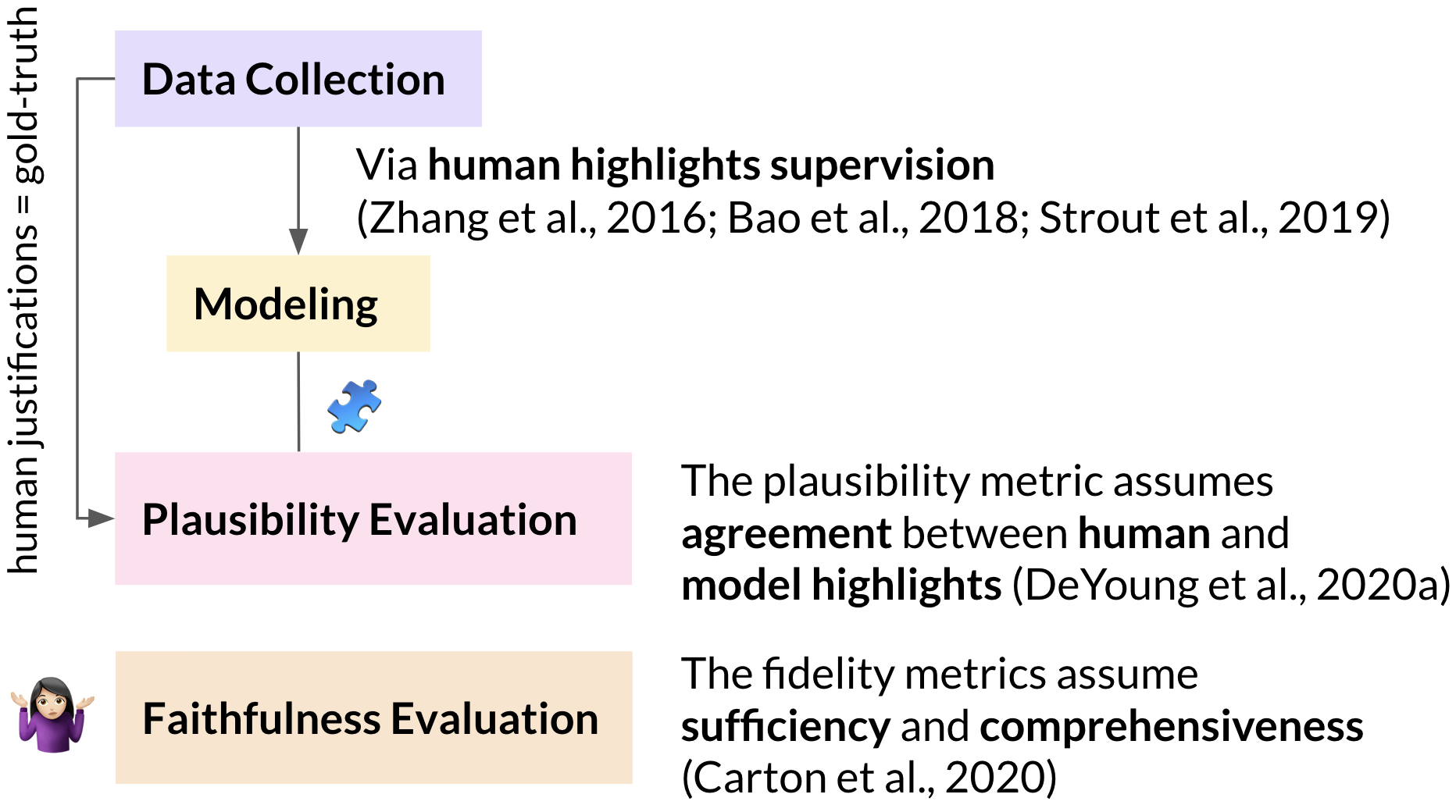}\label{fig:reconcile_1}}
    \qquad
    \par\bigskip
    \subfloat[Unsupervised models' development. In \sect{sec:collecting_highlights}, we illustrate that comprehensiveness is not a necessary property of human highlights.  Non-comprehensiveness, however, hinders evaluating plausibility of model highlights produced in this setting since model and human highlights do not match by design.]{\includegraphics[width=0.7\columnwidth]{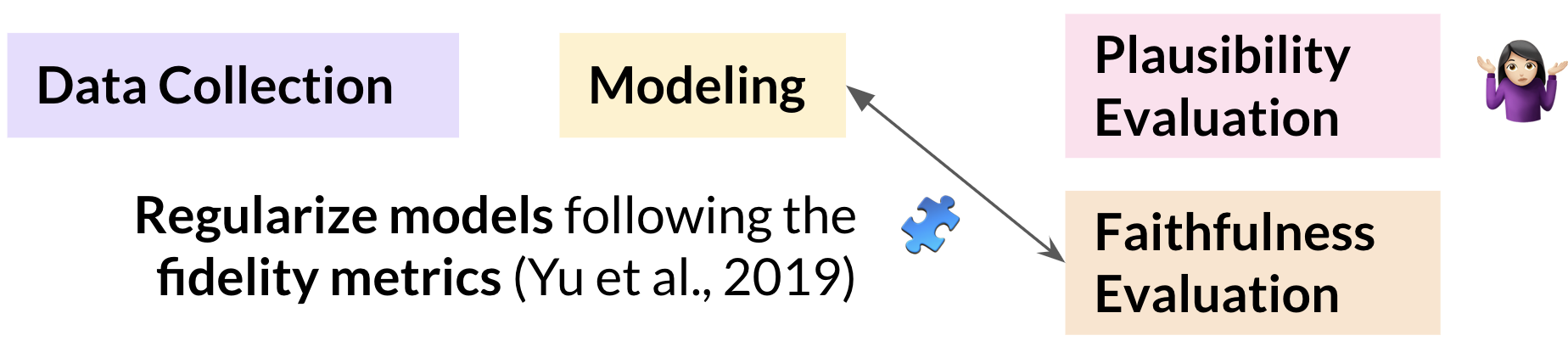} \label{fig:reconcile_2} }
    \qquad
    \par\bigskip
    \centering
    \subfloat[Recommended unsupervised models' development. To evaluate both plausibility and faithfulness, we should collect comprehensive human highlights, assuming that they are already sufficient (a necessary property).]{\includegraphics[width=0.7\columnwidth]{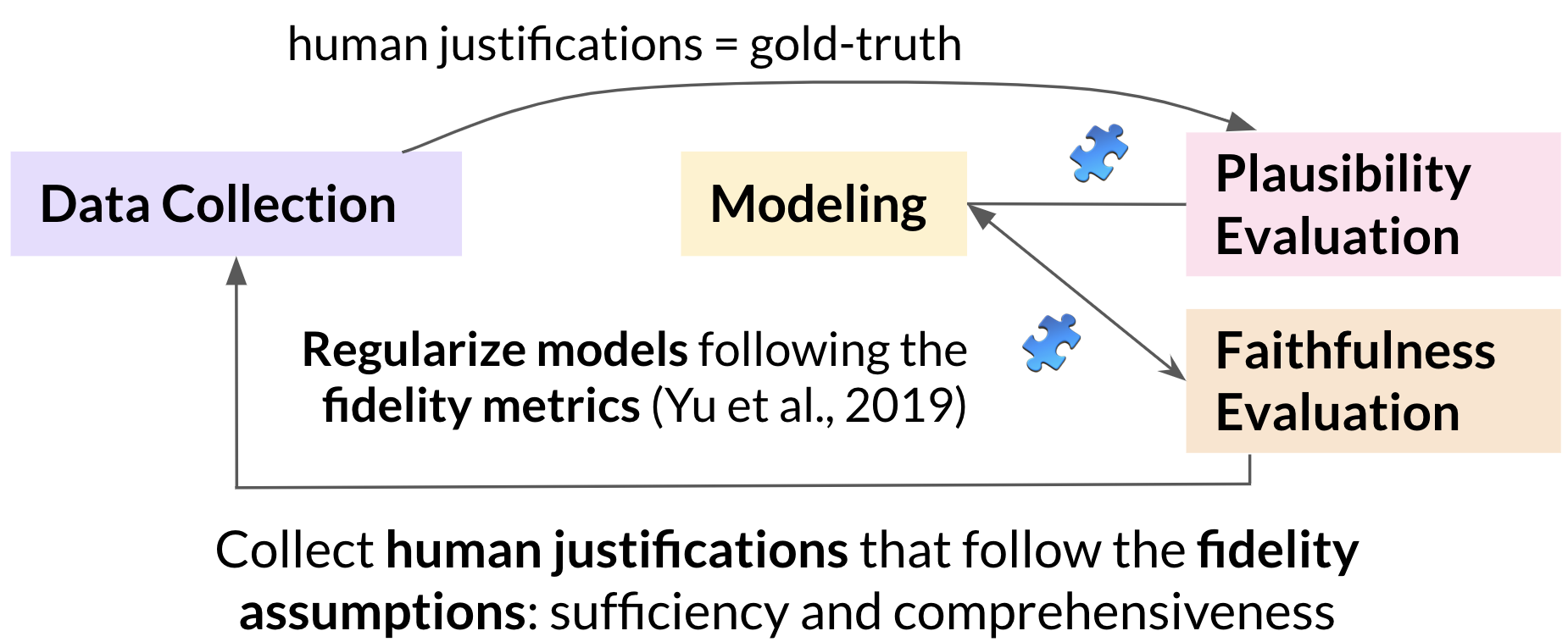} \label{fig:reconcile_3} }
    \par\bigskip
    \caption{Connections between assumptions made in the development of self-explanatory \textbf{highlighting} models. The jigsaw icon marks a synergy of modeling and evaluation assumptions. The arrow notes the direction of influence. The text next to the plausibility / faithfulness boxes in the top figure hold for the other figures, but are omitted due to space limits. Cited: \citet{deyoung-etal-2020-eraser, zhang-etal-2016-rationale, bao-etal-2018-deriving, strout-etal-2019-human, carton-etal-2020-evaluating, yu-etal-2019-rethinking}.}
    \label{fig:reconcile}
\end{figure}

\end{document}